# Demystifying the Potential of ChatGPT-4 Vision for Construction Progress Monitoring

**A. B. Ersoz**
*Middle East Technical University, Civil Engineering Department, Ankara, Türkiye*
abersoz@metu.edu.tr

## Abstract

*The integration of Large Vision-Language Models (LVLMs) such as OpenAI's GPT-4 Vision into various sectors has marked a significant evolution in the field of artificial intelligence, particularly in the analysis and interpretation of visual data. This paper explores the practical application of GPT-4 Vision in the construction industry, focusing on its capabilities in monitoring and tracking the progress of construction projects. Utilizing high-resolution aerial imagery of construction sites, the study examines how GPT-4 Vision performs detailed scene analysis and tracks developmental changes over time. The findings demonstrate that while GPT-4 Vision is proficient in identifying construction stages, materials, and machinery, it faces challenges with precise object localization and segmentation. Despite these limitations, the potential for future advancements in this technology is considerable. This research not only highlights the current state and opportunities of using LVLMs in construction but also discusses future directions for enhancing the model's utility through domain-specific training and integration with other computer vision techniques and digital twins.*

## Introduction

The rapid advancement of Large Language Models (LLMs) has significantly shaped the field of natural language processing, placing these models at the center of text analysis and generation efforts. LLMs, known for their training on large datasets, have developed the remarkable capability to understand and produce text in a way that closely mimics human language. The widespread availability of platforms like ChatGPT (OpenAI, 2022) and Gemini (Google, 2023) has made these technologies accessible to more people, expanding their use across different fields. Alongside the advancements in language processing, the emergence of Large Vision-Language Models (LVLMs), also sometimes referred to as Large Vision Models, signifies a similar revolution within the realm of image processing and interpretation. These sophisticated artificial intelligence systems are engineered to analyze and interpret visual information like their linguistic counterparts interpret text. Drawing on extensive visual datasets, LVLMs are adept at identifying and understanding visual elements and structures, thereby enabling a wide array of applications, including image-based dialogue, image recognition, visual question answering, document analysis, and image captioning.

The ascent of LVLMs follows the initial breakthroughs made by LLMs between 2017 and 2024, as illustrated in Figure 1. One of the first releases of LVLMs with pioneering models is CLIP, released by OpenAI (2021). CLIP, a vision-language model, excels in tasks ranging from image captioning to visual question answering and image retrieval by leveraging its dual understanding of images and text. The advent of GPT-4 (Generative Pre-trained Transformers) by OpenAI marked a significant milestone with its enhanced generative capabilities (OpenAI, 2023a). In September 2023, OpenAI expanded these capabilities by introducing multi-modal functions to ChatGPT titled GPT-4 Vision (GPT-4V), allowing the model to process and respond to visual and auditory data (OpenAI, 2023b). GPT-4V extends the robust capabilities of GPT-4 to include visual analysis, offering a more holistic interaction experience. This multi-modal integration provides new opportunities for various industries to harness AI in unprecedented ways.

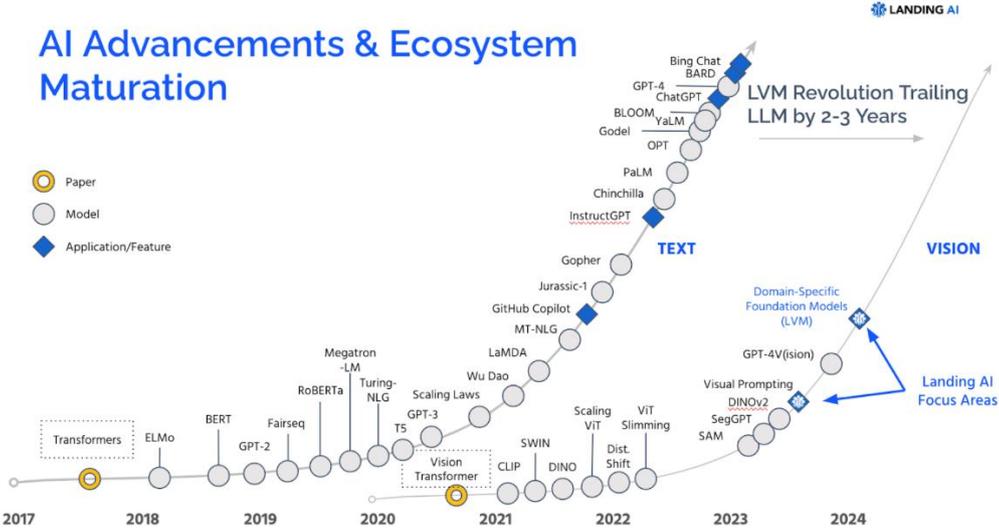

Figure 1 - LVM and LVLM progress (LandingAI, 2023)

Applications of GPT-4 Vision span numerous sectors, including healthcare, robotics, security and surveillance, retail and commerce, as well as content creation and entertainment, demonstrating its versatility and broad utility. This introductory exploration sets the stage for a detailed investigation into the applications of GPT-4 Vision, focusing on the construction industry—a field where the potential of such technologies has not yet been fully tapped. This study aims to reveal the applications of GPT-4 Vision in construction, highlighting its potential in progress monitoring and tracking. As this topic is delved into more deeply, the capabilities and challenges of integrating AI in construction will be explored, providing a comprehensive overview of how these technological advancements could reshape the industry.

## Related Work

The integration of artificial intelligence in various sectors has reached new heights with the advent of GPT-4V, OpenAI's most advanced generative model. This section delves into the diverse applications of GPT-4V across different domains, demonstrating its capability to interpret and respond to visual data alongside text. From transforming medical diagnostics to advancing remote sensing technologies, the applications of this AI tool exemplify a significant leap in bridging the gap between visual perception and language understanding.

In a 2024 study by Sievert et al., the researchers explored how OpenAI's ChatGPT 4 with Vision capabilities could be used to analyze medical images, specifically confocal laser endomicroscopy (CLE) images, for diagnosing a type of cancer that occurs in the throat. A collection of 12,809 CLE images from five patients was gathered and anonymized. The images were used to train and validate OpenAI's GPT-4.0 API. A training dataset of 16 images and a validation dataset of 139 images were classified. Despite the limited training, GPT-4.0 achieved an accuracy of 71.2% in identifying cancer, which was still below the 88.5% accuracy rate of human medical experts. This discrepancy highlights the challenges and limitations of AI in medical diagnostics, mainly when dealing with small datasets (Sievert et al., 2024). In a correspondence to the editor of the International Journal of Surgery (London, England), the potential of GPT-4 in altering breast cancer treatment is highlighted. The authors discuss how this advanced AI technology can aid in interpreting complex medical concepts, providing emotional support, and assisting in patient self-management, making it a critical component of breast cancer care. While GPT-4 enhances decision-making and medical education for healthcare providers, it is emphasized that it cannot replace the crucial role of human doctors or the empathy inherent in patient care. Breast cancer treatment still demands expert judgment and a personalized approach from a physician, and ethical considerations are vital to ensure fairness and transparency in AI's application. Despite potential challenges, there is significant optimism about the role of GPT-4 in advancing breast cancer treatment, promising a future where human expertise and artificial intelligence synergize to enhance patient support and care. The correspondence calls for further research to explore the benefits of GPT-4, underscoring its potential to transform healthcare practices (Deng et al., 2023).

Recent advancements in AI technology have led to exploring its potential applications in medical education, particularly in examining how models like ChatGPT perform in medical examinations. In a 2024 study by Nakao et al., researchers assessed GPT-4V using the 117th Japanese National Medical Licensing Examination, which included visual elements. The study revealed that the inclusion of images did not improve performance; accuracy rates were 68% with images and 72% without, suggesting that the current version of GPT-4V may struggle with interpreting medical images, potentially limiting its applicability in medical training and decision-making (Nakao et al., 2024). Building on these findings, another study by Hyungjin Kim et al. in 2024 explicitly focused on ChatGPT's ability to interpret radiological images from medical school exam questions. While ChatGPT demonstrated proficiency in handling text-based radiology questions, it weakened with image-based queries, showing significantly lower accuracy than third-year medical students. This indicates that although AI models like ChatGPT are progressing, they still require significant enhancements and specialized training in medical imaging to be considered reliable in the clinical radiology setting (Kim et al., 2024).

Despite this technology's challenges and early development stage, the authors are optimistic about its application prospects, suggesting that Visual ChatGPT could significantly advance remote sensing methodologies by making them more accessible and efficient. Balado and Nguyen (2023) evaluated the functionality of MATLAB codes generated by ChatGPT for point cloud processing tasks such as surface normal calculation, curvature, eigenvalues, and eigenfeatures. The study demonstrated that while ChatGPT could generate simple and short functional codes, it struggled with more complex code that required the correct use of loops, indexes, or equations. This research highlights the potential of LLMs in generating code for specific tasks, suggesting their utility primarily in educational contexts or as a starting point for more complex programming challenges.

ChatGPT-4's vision capabilities have not been thoroughly explored in construction applications, with only a few review papers discussing the potential usage of these vision applications. The 2024 article by Ghimire, Kim, and Acharya explores the role of Generative AI in the construction industry, focusing on its opportunities and challenges. While their discussion primarily addresses text-to-text applications, they also briefly mention text-to-image applications. These potential uses involve generating visual representations of site conditions, traffic patterns, and zoning laws and creating targeted safety and trade training materials. However, the article lacks examples of image-to-text applications (Ghimire et al., 2024). In another study, Saka et al. (2024) comprehensively examine GPTs within the construction industry, exploring their potential applications, limitations, and practical utility. They highlight two specific computer vision capabilities of GPTs: hazard identification and waste management. The authors include an image of a hazardous construction site environment to prompt the identification of potential hazards. Another example demonstrated the use of ChatGPT to identify waste materials in an image and recommend appropriate waste management methods, with ChatGPT providing effective responses. Together, these studies reflect the growing interest in applying advanced AI tools to bridge technological gaps and enhance operational efficiencies in the construction industry.

## An Illustrative Example

This section demonstrates the application of GPT-4 Vision to monitor and track construction progress for two buildings on a university campus at the Middle East Technical University. The first site features a combination of reinforced and steel structures (Figure 2a), while the second site consists of fully reinforced concrete structures (Figure 2b). ChatGPT had been used via chat.openai.com as of April 28th, 2024. We used high-resolution aerial photographs with dimensions of 4000x3000 pixels. No custom instructions of ChatGPT or memorization features were enabled, allowing results based purely on the input provided during each session. The response time for each result generation was less than 30 seconds.

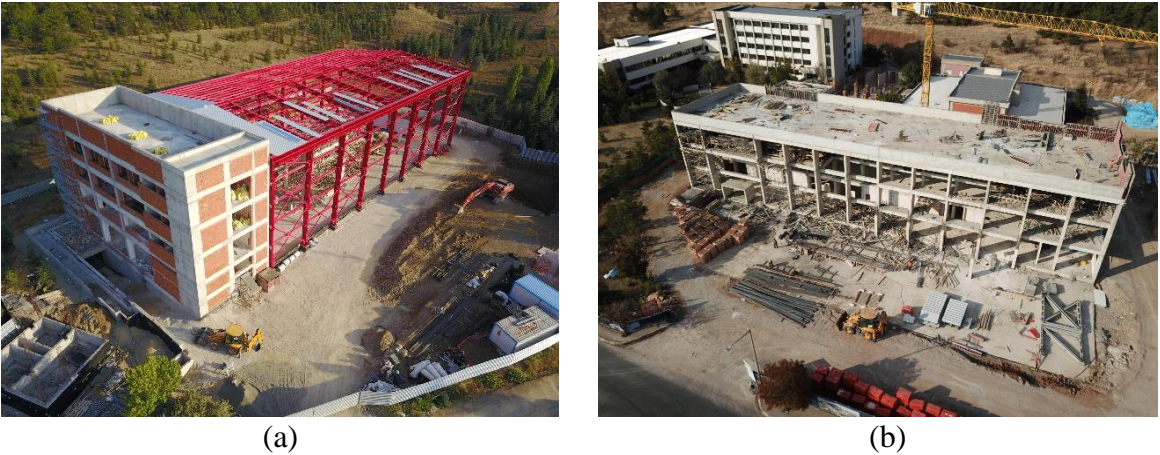

(a)                        (b)

Figure 2 - Input images for scene understanding

For the initial analysis, High-resolution aerial images measuring 4000x3000 pixels were utilized. The AI was prompted to *"List the properties in this construction site."* for each image, ChatGPT responded to for reinforced and steel structure (Figure 2a) as:

> *In the construction site depicted in the image, the following properties and features can be identified:*
> *__1. Building Under Construction__: A partially constructed multistory building with exposed floors.*
> *__2. Red Steel Framework:__ A red steel structural framework is visible on top of the building, indicating ongoing construction or expansion.*
> *__3. Heavy Machinery:__*
> *• Excavator: Positioned on the ground to the right of the building, possibly used for earthmoving or site preparation tasks.*
> *• Wheel Loader or similar machinery: Located at the bottom left corner, typically used for moving materials.*
> *__4. Construction Materials:__*
> *• Sand piles: Likely used for making concrete or for grading purposes.*
> *• Gravel or aggregate piles: Also a component of concrete or used for drainage systems.*
> *• Pipes and construction supplies: Stored around the site, indicating the various stages of construction activity.*
> *__5. Perimeter Fencing:__ Enclosing the site for security and safety, indicating a controlled access area.*
> *__6. Storage and Workspace:__*
> *• Covered areas that could serve as on-site storage for tools and materials or as workspaces for the construction crew.*
> *• Open areas with stacked materials and supplies necessary for the ongoing construction work.*
> *__7. Geographic Location:__ The presence of a forest or treed area around the construction site, suggesting a location that is not densely urban.*
> *__8. Site Terrain:__ Modified ground surrounding the building, with visible tire tracks and dug-out areas indicating heavy machinery movement and site preparation work.*
> *__9. Safety Features__: Visible safety barriers around potential hazards and possibly safety signage (not clearly visible but commonly present in such settings).*

ChatGPT response for fully reinforced concrete structure (Figure 2b):

> *The image shows a construction site with various features:*
> *__1. Partially Constructed Multi-Story Building__: There's a structure that appears to be a multi-story concrete building in the midst of construction. Some floors are partially done with visible columns and support beams.*
> *__2. Construction Materials:__ Various construction materials are scattered throughout the site, including stacks of bricks, piles of construction bars (rebar), and other building materials.*
> *__3. Heavy Machinery:__ There is construction machinery present, which includes at least one telehandler and what appears to be a mobile crane.*
> *__4. Scaffolding:__ There are sections of scaffolding around the building, used by workers to access higher levels of the construction.*
> *__5. Debris:__ There's a significant amount of debris and rubble on the upper floors of the building, possibly from demolition work or the aftermath of a structural collapse.*
> *__6. Surrounding Structures:__ Adjacent to the construction site, there are other buildings that look like completed structures, possibly office or industrial buildings.*
> *__7. Safety Barriers__: There are red and white safety barriers around the perimeter of the site, which are likely there to prevent unauthorized access and to mark the construction zone.*
> *__8. Construction Site Accessories:__ There are also various items like portable toilets, storage containers, and makeshift sheds, probably used for storage or as temporary site offices.*
> *__9. Fenced Areas:__ Part of the construction site is fenced, likely for security purposes and to delineate the construction area from the surrounding environment.*
> *__10. Topography:__ The site is surrounded by some greenery and trees, indicating it may be on the outskirts of an urban area or in a more suburban or rural setting.*
> *__11. Vehicle Tracks:__ There are visible tracks on the ground, indicating the movement of vehicles around the site.*
> *These observations provide a general understanding of the construction site's characteristics and status.*

These results showcase GPT-4 Vision's adeptness at comprehensive scene analysis, providing a foundational understanding of the construction site's current state and context. Following the scene understanding, the study progressed to examine GPT-4 Vision's capability in tracking construction development and assessing task completion. Two sets of aerial images of the fully reinforced concrete structure, taken one month apart, were uploaded to the ChatGPT (Figure 3). The system was instructed to *"Two images will be uploaded, after upload, categorize and create a list of construction progress."*

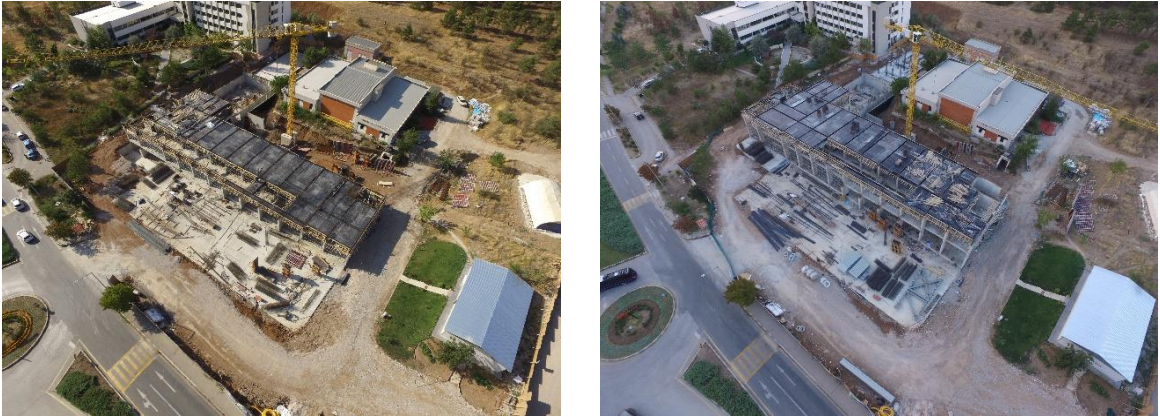

Figure 3 - Input images for construction progress tracking

ChatGPT responded to this prompt as given below:

> *Thank you for uploading both images. Based on the visual information, I will categorize the construction progress as follows:*
> ***First Image:***
> *1. Excavation and foundation work appears to be complete.*
> *2. Formwork for columns and walls is erected, indicating that the construction is prepared for concrete pouring.*
> *3. Scaffolding is in place for work on the above-ground structure.*
> *4. Construction materials and equipment are organized around the site, showing active work. 5. The ground floor slab seems to be complete, with work progressing on upper floor slabs.*
> ***Second Image:***
> *1. The progress is similar to the first image with the continuation of concrete work.*
> *2. Formwork is present on additional floors, suggesting vertical construction progression. 3. More scaffolding is visible, indicating ongoing work on higher levels.*
> *4. Construction materials appear to be more spread out, likely due to work taking place on various parts of the building simultaneously.*
> *5. The perimeter of the building seems to be more defined, with potential work on internal partitions or walls.*
> *Both images show the construction site at an advanced stage of the structural phase, with work focused on erecting the concrete skeleton of the building. The precise sequence of construction steps can only be determined if the images are taken at different times; however, the visible elements suggest the project is moving from completing the ground floor toward constructing the upper floors. There doesn't appear to be any significant exterior work yet, such as cladding or facade installation, which typically follows the completion of the main structure.*

Then, the AI tool was provided with a list of expected tasks at the construction site and asked to, *"From this list which tasks are completed? Categorize, completed, ongoing and not started tasks only. Prepare a table.*
*1 Site - Earth*
*2 Foundations*
*3 Structural Columns - B1 Basement*

*4 Structural framing - B1 Basement*
*5 Floor Slabs - B1 Basement*
*6 Structural Framing - Ground Floor*
*7 Floor Slabs - Ground Floor*
*8 Structural Columns - Ground Floor*
*9 Columns - Ground Floor*
*10 Structural Framing - 1st Floor*
*11 Floor Slabs - 1st Floor*
*12 Structural Columns - 1st Floor*
*13 Columns - 1st Floor*
*14 Structural Framing - 2nd Floor*
*15 Floor Slabs - 2nd Floor*
*16 Structural Columns - 2nd Floor*
*17 Columns - 2nd Floor*
*18 Structural Framing - Roof*
*19 Floor Slabs - Roof*
*20 Roofs"*

GPT-4 Vision identified the progress and tabulated the tasks into categories as below:

> *Based on the images provided and the construction list, the categorization of tasks is as follows:*
> *1 Site - Earth - Completed*
> *2 Foundations - Completed*
> *3 Structural Columns - B1 Basement - Completed*
> *4 Structural framing - B1 Basement - Completed*
> *5 Floor Slabs - B1 Basement - Completed*
> *6 Structural Framing - Ground Floor - Completed*
> *7 Floor Slabs - Ground Floor - Completed*
> *8 Structural Columns - Ground Floor - Completed*
> *9 Columns - Ground Floor - Completed*
> *10 Structural Framing - 1st Floor – Ongoing*
> *11 Floor Slabs - 1st Floor – Ongoing*
> *12 Structural Columns - 1st Floor – Ongoing*
> *13 Columns - 1st Floor – Ongoing*
> *14 Structural Framing - 2nd Floor – Not Started*
> *15 Floor Slabs - 2nd Floor – Not Started*
> *16 Structural Columns - 2nd Floor – Not Started*
> *17 Columns - 2nd Floor – Not Started*
> *18 Structural Framing - Roof – Not Started*
> *19 Floor Slabs - Roof – Not Started*
> *20 Roofs – Not Started*
> *The above categorization is inferred from the images with the understanding that the first floor is above the ground floor. The tasks that are 'Ongoing' are assumed because there is visible work in progress related to those tasks in the images, whereas 'Not Started' tasks are those for which there is no visible work or structure present.*

## Discussion and Conclusions

The deployment of GPT-4 Vision in construction progress monitoring offers promising insights into the initial capabilities of AI for scene understanding and progress monitoring. As far as the authors' knowledge, this is the first study to explore the use of GPT-4 Vision for construction progress monitoring, offering new insights into its capabilities and limitations. This section delves into the findings of this evaluation in detail while explaining the current state, limitations, opportunities, and future works.

**Current State and Limitations**

- The review of ChatGPT-4 Vision capabilities in identifying and analyzing construction sites reveals both its strengths and limitations. Initially, ChatGPT effectively recognizes the type of building and construction materials, correctly identifying type of steel and reinforced concrete structures.
- Regarding construction machinery, ChatGPT demonstrates an understanding of different equipment types, such as excavators and wheel loaders, although it struggles with precise location identification due to its inability to detect objects at the pixel level. The inability to accurately detect objects in images hinders the performance of detection and segmentation tasks in computer vision applications. The machinery was identified as a telehandler and a mobile crane in Figure 2b. Although the mobile crane was correctly identified, a backhoe loader was actually present on the site instead of a telehandler. However, the equipment mounted on the front of the backhoe loader is similar to that of a telehandler, which performs similar functions. Therefore, this misclassification is not entirely incorrect.
- Moreover, ChatGPT can identify various construction site materials like sand piles, pipes, and other supplies, providing relevant information about their uses. This knowledge aids in assessing the progress of construction accurately. However, there are instances of misclassification, such as wrongly identifying debris on upper floors as unused materials, leading to incorrect assumptions about potential demolition work. This points to limitations in object classification that could affect the model's reliability in specific scenarios.
- Regarding safety measures, ChatGPT identifies general precautions like perimeter fencing and safety barriers but lacks specificity about their locations. This general information might not be sufficient for detailed safety analysis or planning.
- ChatGPT demonstrates proficiency in contextualizing the construction environment, successfully identifying non-urban settings by noting surrounding forests and tree areas. It recognizes site accessories such as portable toilets and storage containers, which are crucial for managing and organizing construction sites. The model also comments on site accessibility by noting roads and visible tracks, although these observations are not detailed.
- For the comparison of consecutive time images, ChatGPT analyzes each construction stage sequentially, providing general information about the stage, materials, and machinery involved as it conducts scene understanding (Figure 3). The first image was identified with the foundation work completed. ChatGPT noted the preparation for concrete pouring, inferred from the erection of formwork for columns and walls. It correctly identified the completion of the ground floor slab. The addition of floors and the progression of vertical construction was recognized in the second image. It was also observed that more construction materials were distributed across the site, indicating that various parts of the structure were being built simultaneously.
- To establish a baseline for ChatGPT, items from the construction plan were provided, and it was prompted to categorize these items as completed, ongoing, or not started. It demonstrated the ability to create a categorized table. ChatGPT accurately understood that the earthworks and foundations were completed. The platform identified the floor levels, noting that the tasks on the ground floor were completed and that construction on the first floor was ongoing. The tasks for the second floor had not yet started. A more detailed construction plan could be input to enhance the categorization of each task.

**Opportunities and Future Works**

- Looking forward, the development of GPT-4V in progress monitoring tasks represents an early step with significant potential for expansion and improvement. Current research should include more diverse structures, such as transportation and hydraulic facilities, to test the model's capability to identify different material types. This would provide a broader understanding of its applicability across various construction domains.
- Integrating aerial and ground images when uploading to the ChatGPT process could enhance the model's effectiveness by providing a more comprehensive view of construction sites. This approach could lead to more detailed and accurate monitoring, encompassing a wider range of perspectives and details.
- While current GPT models do not have the capability to segment objects within images, utilizing pre-segmented images could facilitate better object differentiation, as suggested by Yang et al. (2023), indicating that segmented images improve object localization. Hence, incorporating segmented images could be a valuable next step in refining GPT's visual recognition functions.
- It is also crucial to evaluate other language vision models like Gemini (Google, 2023) and LLaVa (Liu et al., 2023) to compare their performance with GPT-4V. Such comparative studies could highlight strengths and weaknesses specific to each model, guiding further development and optimization.
- Integrating GPT models with as-planned models could significantly enhance construction monitoring, such as exporting screenshots of the same viewpoints from 4D BIM as both as-planned and as-built images for GPT analysis. This integration could assist in creating accurate digital twins and enable precise tracking of construction progress.
- Lastly, there is immense potential in training or fine-tuning these models with construction-specific images. Domain-specific large vision models could revolutionize the industry by improving efficiency and expanding capabilities (LandingAI, 2023). Their ability to understand complex patterns from large datasets could be pivotal in addressing various challenges within the construction sector and beyond.

https://blog.google/technology/ai/google-gemini-ai/

Kim, H., Kim, P., Joo, I., Kim, J. H., Park, C. M., & Yoon, S. H. (2024). ChatGPT Vision for Radiological Interpretation: An Investigation Using Medical School Radiology Examinations. *Korean Journal of Radiology*, *25*(4), 403–406. https://doi.org/10.3348/kjr.2024.0017

LandingAI. (2023). *Introducing Domain-Specific Large Vision Models*. https://landing.ai/blog/introducing-domain-specific-large-vision-models

Liu, H., Li, C., Wu, Q., & Lee, Y. J. (2023). Visual Instruction Tuning. *ArXiv Preprint*.

Nakao, T., Miki, S., Nakamura, Y., Kikuchi, T., Nomura, Y., Hanaoka, S., Yoshikawa, T., & Abe, O. (2024). Capability of GPT-4V(ision) in the Japanese National Medical Licensing Examination: Evaluation Study. *JMIR Medical Education*, *10*, e54393. https://doi.org/10.2196/54393

OpenAI. (2021). *CLIP: Connecting text and images*. https://openai.com/research/clip

OpenAI. (2022). *Introducing ChatGPT*. https://openai.com/blog/chatgpt

OpenAI. (2023a). *GPT-4 is OpenAI's most advanced system, producing safer and more useful responses*. https://openai.com/gpt-4

OpenAI. (2023b). *GPT-4V(ision) system card*. https://openai.com/research/gpt-4v-system-card

Saka, A., Taiwo, R., Saka, N., Salami, B. A., Ajayi, S., Akande, K., & Kazemi, H. (2024). GPT models in construction industry: Opportunities, limitations, and a use case validation. *Developments in the Built Environment*, *17*(December 2023), 100300. https://doi.org/10.1016/j.dibe.2023.100300

Sievert, M., Aubreville, M., Mueller, S. K., Eckstein, M., Breininger, K., Iro, H., & Goncalves, M. (2024). Diagnosis of malignancy in oropharyngeal confocal laser endomicroscopy using GPT 4.0 with vision. *European Archives of Oto-Rhino-Laryngology*, *281*(4), 2115–2122. https://doi.org/10.1007/s00405-024-08476-5

Yang, J., Zhang, H., Li, F., Zou, X., Li, C., & Gao, J. (2023). Set-of-Mark Prompting Unleashes Extraordinary Visual Grounding in GPT-4V. *ArXiv Preprint*, 1–23.